# Analyzing Periodicity and Saliency for Adult Video Detection

Yizhi Liu[1], Xiaoyan Gu[2,*], Lei Huang[3], Junlin Ouyang[1], Miao Liao[1], Liangran Wu[1]

[1]College of Computer Science and Engineering, Hunan University of Science and Technology, Xiangtan, China, 411201

[2]Institute of Information Engineering, Chinese Academy of Sciences, Beijing, China, 100093

[3]College of Information Science and Engineering, Ocean University of China, Shandong, China, 266100

*corresponding author: guxiaoyan@iie.ac.cn

Abstract: Content-based adult video detection plays an important role in preventing pornography. However, existing methods usually rely on single modality and seldom focus on multi-modality semantics representation. Addressing at this problem, we put forward an approach of analyzing periodicity and saliency for adult video detection. At first, periodic patterns and salient regions are respectively analyzed in audio-frames and visual-frames. Next, the multi-modal co-occurrence semantics is described by combining audio periodicity with visual saliency. Moreover, the performance of our approach is evaluated step by step. Experimental results show that our approach obviously outperforms some state-of-the-art methods.

## 1 Introduction

With the explosive growth of online videos, filtering pornography has growing crucial. The precise definition of pornography is subjective, but here we will consider "any sexually explicit material with the aim of sexual arousal or fantasy" [28]. Content-based adult video detection is one of the most powerful approaches.

The traditional methods are based on visual features of keyframes (namely images). In terms of the areas from which visual features are extracted, it can be classified into three kinds: global-features based methods [33-36], bag-of-words based methods [37-40], and region-of-interest (ROI) based methods [41-44]. However, it is difficult for the keyframe-based approach to detect adult videos accurately owing to the challenges of detecting salient regions [1-12], extracting spatial-temporal features [13-19], and representing multimodal co-occurrence semantics.

Salient object detection models usually aim to detect only the most salient objects in a scene and segment the whole extent of those objects [1-10]. In computer vision, it is usually interpreted as a process that includes two stages [1]: 1) detecting the most salient object and 2) segmenting the accurate region of that object. Ali Borji et al. [1] proposed that a good saliency detection model should cover "good detection" and "high resolution" at least. In the field of adult image detection, salient object means pornographic parts with skin-color. However, skin-color regions are always reckoned as region-of-interest (ROI) which is actually larger than sub-areas containing pornographic parts.

Audio signal is a conspicuous modality of adult videos. The obvious temporal information in adult videos is the characteristic of periodic patterns, such as periodic moaning and screaming [20]. Zuo et al. [21] proposes a method in which audio waveforms are transformed into a sequence of feature vectors and classified by a Gaussian mixture model. Nevertheless, low-level audio features are too similar to be distinguished accurately.

Adult video detection is intrinsically a multimodal problem since both audio and visual signals can provide important clues. These clues appeared simultaneously and are hidden in the audio periodicity and the visual saliency. Therefore, we are motivated by three aspects: audio periodicity analysis, visual saliency analysis, and multimodal co-occurrence semantic representation.

In this paper, we propose a novel approach of analyzing periodicity and saliency for adult video detection. Its technical contribution lies in:
- The multi-modal co-occurrence semantics is described by combining audio periodicity with visual saliency.
- We successively propose two periodic algorithms echoing with each other. One is the periodic audio codebook algorithm, and the other is the periodic video decision algorithm.
- Saliency based visual codebook algorithm is presented after discovering a hybrid ROI detection model on the basis of saliency and skin-color models.
- Experimental results show that our approach is able to accurately detect adult videos and obviously outperforms some state-of-the-art methods.

The rest of this paper is organized as follows. Section 2 includes related work. In section 3 we introduce the framework and the details of our approach. The experiments described in section 4, and the conclusions are finally given in the last section.

## 2 Related Works

Adult video detection [20-29] can be reckoned as a particular kind of video event detection [13-19]. It is closely related to adult image detection [33-44], salient object detection [1-12], and audio content analysis [30-32].

Content-based adult video detection is traditionally based on visual features of keyframes. Forsyth et al. [33] construct a human figure grouper after detecting skin regions, but consuming a lot of time and low detection accuracy are the two shortcomings. Zeng et al. [34] implement the image guarder system to detect adult images by different kinds of global features. Rowley et al. [35] adopt 27 visual features for Google to filter adult images, including color, shape, texture, face and etc. Tang et al. [36] employ latent Dirichlet allocation to cluster images into some subsets and then combine supported vector machines (SVM) classifiers on one subset to rate adult images. Jones and Rehg [41] developed a statistical color model for detecting skin regions. AIRS [42] employs the MPEG-7 visual descriptors to identify and rate adult images. Kuan and Hsieh [43] used image retrieval technique and visual features are extracted from skin regions. Zheng et al. [44] used an edge-based Zernike moment method to detect harmful symbol objects. However, it is difficult to detect accurately the images whose background colors are similar to skin-color. And it is challenging to identify images in the presence of occlusion, background clutter, pose and lighting changes.

ROI-based methods are more accurate in describing the image content than global-feature based methods for the scenes involving a clutter of prominent and/or specific objects against background and/or foreground [1]. ROI plays an important role in adult image detection. Traditionally, skin-color regions are always extracted as ROI. Lee et al. [22] used a linear-discriminative classifier to combine two frame-based methods, one using on a skin probability map, the other color histograms. Kim et al. [23] used a shape description of skin areas in video frames. A manually defined color range is used for deciding whether a pixel belongs to a skin area. The area's shape is then described by normalized central moments and matched to samples in a database. However, skin-color regions are always larger than the sub-areas containing pornographic parts, and the approach is difficult to differentiate between human skins and other objects with the skin-colors.

Visual attention is a mechanism which filters out redundant visual information and detects the most relevant parts of our visual field [9]. Attention is a general concept covering all factors that influence selection mechanisms, whether they be scene-driven bottom-up or expectation-driven top-down [10, 11]. Therefore, we are motivated to integrate visual attention models with skin-color models [52] and face detection models [53], and to further devise a hybrid approach of ROI detection.

Adult image detection based on codebooks (or bag-of-words) has been a promising approach [37-40]. Deselaers et al. [37] combine it with color histogram to classify images into different categories of pornography. Wang et al. explored an algorithm to reduce the size of visual codebook and to integrate it with spatial distribution [38]. SURF [54] and ROI based visual codebook are respectively applied for adult image detection [40]. Some other approaches [45-53] are worthy to be noticed.

Other modalities have been employed for adult video content classification. The periodicity in the motion information of adult videos is firstly aroused the attention [24, 25]. Audio information is another discriminative clue for adult video detection. Rea et al. [20] combined skin color estimation with the detection of periodic patterns in a video's audio signal. The literatures [30-32] divide audio signals into audio segments to improve the distinguishable power by increasing granularity.

## 3 The Framework and Details of Our Approach

As Figure 1 shown, the framework of our approach includes six main modules: audio segmentation, visual keyframe selection, periodicity analysis based feature extraction, saliency analysis based feature extraction, multi-modal co-occurrence semantic feature representation, and periodicity based classification.

In our framework, audio frames are segmented into units of energy envelope (EE) on the basis of periodicity analysis, and visual keyframes are detected based on saliency analysis. The EE's lengths are not the same but variable. Subsequently, audio signals are depicted via the EE sequences with audio codebook based representation. Visual codebook is generated in light of our ROI model, which combines saliency analysis with a skin-color model. Results show that our framework achieves more excellent performance than some state-of-the-art methods.

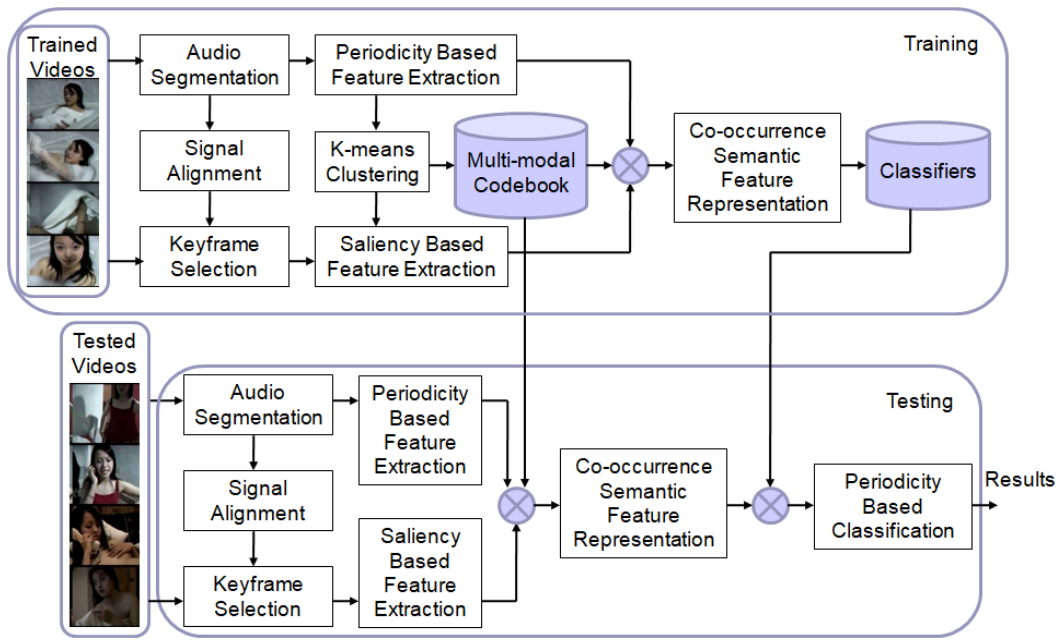

**Fig.1. The framework of our approach**

### 3.1 Periodicity analysis based feature extraction

Some parameters of audio data are set as follows: mono-channel, 16 bits resolution, and the sample rate is 44100 Hz. We extract 36-dimentional feature vectors as an audio frame with 20 milliseconds intervals and no overlapping: 13-dimentional mel-frequency cepstral coefficients (MFCC), 13-dimentional MFCC differential coefficients, 1-dimentional zero crossing rate, 1-dimentional short time energy, 4-dimentional sub-band short time energy, and 4-dimentional sub-band short time energy ratio.

In order to generate audio codebook more accurately, we select some the conspicuous adult videos and segment the noticeable parts as the training dataset. Audio signals are segmented into some EE sequences. Visual keyframe is chosen in the time window of each EE. Figure 2 shows the alignment method of the two signals.

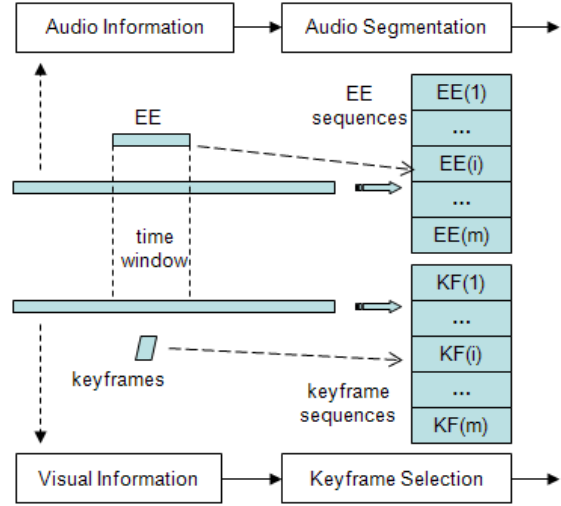

**Fig.2. The alignment method of visual and audio signals**

Periodicity based audio codebook algorithm is consisted of two phases: audio segmentation based on the audio periodicity, codebook generation and middle-level feature representation. Audio features are extracted from audio files. And they are smoothed by the following detection function (1).

$$d_i = \max_{j=1,...,J}(\frac{E_{i+j}}{E_i}) \quad (1)$$

where $d_i$ and $E_i$ mean the value of detection function and the short time energy for the $i-th$ frame respectively. The segmentation positions of EE sequences are computed via the detection function (2) and the thresholds $T_1$ and $T_2$. If the EE length is less than the maximal length $L_{max}$, it can be divided into groups of audio frames. Otherwise, an EE sequence will be segmented into smaller EE sequences according to their periodicity $T$.

$$P(i) = \begin{cases} 1, & d_i \geq T_2 ; \\ \frac{d_i - T_1}{T_2 - T_1}, & T_1 < d_i < T_2 ; \\ 0, & d_i \leq T_1 . \end{cases} \quad (2)$$

After dividing the EE sequences into some manually-selected adult videos, audio codebook is generated via the K-means clustering algorithm. The low-level features of audio frames are respectively compared with each element of audio codebook, and then vote to that with the least Euclidean distance. The votes to the audio codebook are divided by the length of each EE so as to lessen the negative effect of different lengths. Thus, these quotients reflect the occurrence times of audio codebook, which imply some certain middle-level semantics.

## 3.2 Saliency analysis based feature extraction

There are two typical visual attention models, the saliency-based model [10] and the contrast-based model [11]. However, the previous model is time-consuming due to massive computations and the latter one is limited to highlight human-beings in the images. Therefore, we propose the algorithm to fuse the two preceding models. Also, the hybrid ROI detection method consists of three models: saliency analysis model, skin-color model and face detection model, as shown in Figure 3. A visual keyframe is provided as an input image and its ROI $S$ is distilled via the hybrid ROI detection algorithm.

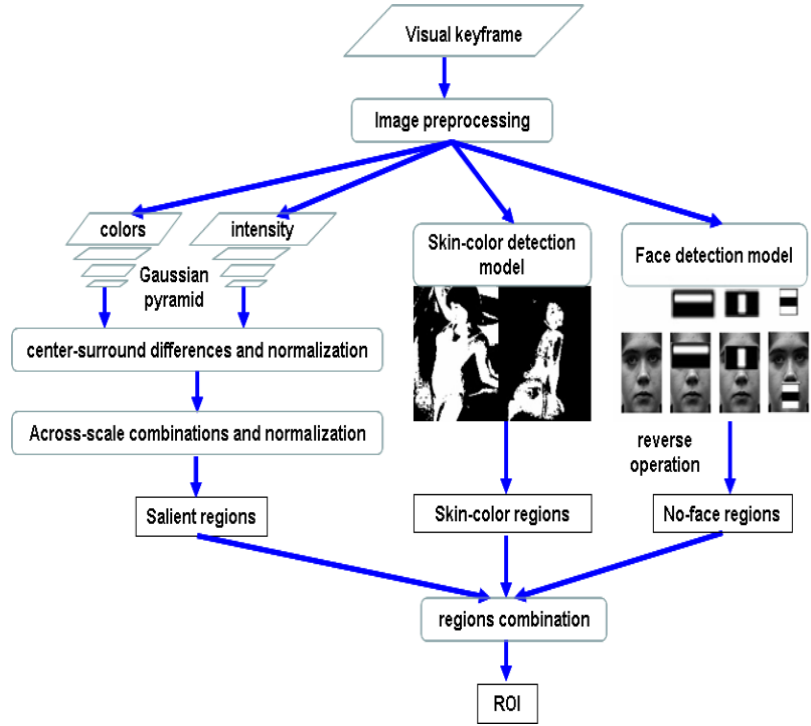

**Fig.3. The hybrid approach of ROI detection algorithm.**

The first step is image preprocessing. The input keyframe is resized to a uniform image without changing its appearance ratio. To reduce efficiently the computational complexity, it is quantized to $m \times n$ image blocks. Besides, we arbitrarily set $m = 640$ and $n = 480$ here.

Secondly, multi-scales and contrasts are devised to analyze each image block. The image block $(i, j)$ is reckoned as the perceivable unit $p_{i,j}$ ($i \in (0, m)$ and $j \in (0, n)$). According to the formula（3）, we can calculate the contrast value of its neighborhood $C_{i,j}$, where $\theta$ denotes the neighborhood of $p_{i,j}$; $q$ is one of the signals in $\theta$; $d$ represents the distance between $p_{i,j}$ and $q$.

$$C_{i,j} = \sum_{q \in \theta} d(p_{i,j}, q) \qquad (3)$$

Next, An intensity image $I$ is obtained by $I = (r+g+b)/3$, where r, g, and b represent the red, green, and blue channels of the input image. Then $I$ is used to create a Gaussian pyramid $I(\sigma)$, where $\sigma \in [0,8]$ is the scale. After that, four Gaussian pyramids $R(\sigma)$, $G(\sigma)$, $B(\sigma)$, and $Y(\sigma)$ are built from the broadly-tuned color channels, that is, $R = r - (g+b)/2$ represents red, $G = g - (r+b)/2$ stands for green, $B = b - (r+g)/2$ is blue, and $Y = (r+g)/2 - |r-g|/2 - b$ for yellow (negative values are set to zero).

To create three sets of feature maps, each is computed by a set of linear "center-surround" operations, denoted as $\Theta$. The feature maps are computed followed the below equations (4) to (6), which include the intensity contrast $I(c,s)$, red/green and green/red double opponency $RG(c,s)$ and blue/yellow and yellow/blue opponency $BY(c,s)$ (where $c \in \{2,3,4\}$, $s = c + \delta$, $\delta = \{3,4\}$).

$$I(c,s) = |I(c)\Theta I(s)| \qquad (4)$$

$$RG(c,s) = |(R(c) - G(c))\Theta(G(s) - R(s))| \qquad (5)$$

$$BY(c,s) = |(B(c) - Y(c))\Theta(Y(s) - B(s))| \qquad (6)$$

At last, the saliency map is obtained by the final input $S$, which consists of two normalized and summed conspicuity maps. To obtain the two "conspicuity maps", we used the set $I(c,s)$ for "intensity conspicuity maps $\bar{I}$" via equation (7). Then the latter two sets of $RG(c,s)$ and $BY(c,s)$ are applied for "color conspicuity maps $\bar{C}$" via equation (8). The normalization operation is denoted as $N(\cdot)$ in equation (9), and $C_{i,j}$ is normalized into [0, 255] in each image block.

$$\bar{I} = \bigoplus_{c=2}^{4} \bigoplus_{s=c+3}^{c+4} N(I(c,s)) \qquad (7)$$

$$\bar{C} = \bigoplus_{c=2}^{4} \bigoplus_{s=c+3}^{c+4} [N(RG(c,s)) + N(BY(c,s))] \qquad (8)$$

$$S = \frac{1}{2}(N(\bar{I}) + N(\bar{C})) \qquad (9)$$

Then we adopt the skin-color model proposed by Garcia et al. [55] and the face detection model proposed by Viola [56]. We take the intersecting part of salient regions, skin-color regions, and no-face regions as ROI. To reduce background noises, visual codebook is created from the patches in ROI. Moreover, SURF is adopted because it is more excellent than the state-of-the-arts both in speed and accuracy [40].

### 3.3 Co-occurrence semantic representation and multi-modal fusion

Multi-modal co-occurrence semantic is described by combining audio codebook with visual codebook, as shown in Figure 1. At first, we select some adult videos and extract audio and visual low-level features. After audio periodicity analysis and ROI detection, the audio codebook and the visual codebook are respectively created by K-means clustering algorithm. Next, low-level audio and visual features of the testing video are respectively converted into mid-level semantic histograms via the audio or visual codebook. The histograms are concatenated to represent the co-occurrence semantics of multi-modal (audio and visual) signals. Finally, we adopt periodicity based video decision algorithm to fuse the classification results of multi-modal codebooks with that of visual global features.

Periodicity based video decision algorithm is used to judge that a video is pornographic or not. Traditionally, it is determined by the number of pornographic keyframes in a video. When the number is greater than certain threshold, the video is considered to be pornographic. Otherwise, the video is considered to be innocuous. It is called the video decision algorithm using thresholds in the following. In this paper, we divide the audio stream of a video into EE sequences based on audio periodicity basis. If pornographic EEs consecutively occur and the number of consecutive counter outnumbers the threshold N, the video should be regarded as pornography. In the periodic video decision algorithm, we provide the SVM prediction file as "InputFile", and regulate the parameters $Thr$ and $N$ as the threshold. At first, some variables are initialized: $bool\ flag = false$, $counter = 0$, $firstV = 0$, $secondV = 0$. Then, it reads the first value of InputFile $s$. If $s > Thr$, we set $firstV = 1$, otherwise, we set $firstV = -1$.

Then, full of InputFile is obtained to get the next value $t$ until the end of the file according to the concessive occurrence times. Finally, the decision result $flag$ is returned as the output. If the value is true, the testing video is regarded as pornographic.

## 4 Experiments

We collect videos from the Internet and respectively set up a training dataset and a testing dataset. There are forty eight adult videos and three hundred benign ones in the training dataset. And the testing dataset includes fifty adult videos and one hundred and fifty benign ones. We evaluate our approach in the visual studio 2003 environment with the machine of 1.86 GHz Duo CPU and 2GB memory. On the basis of our previous works [36, 39, 40], we adopt color moments as the global features, SURF as the local features, and SVM classifiers. We evaluate our method with receiver operating characteristic (ROC) curves. A ROC space is defined by false positive rate (FPR) and true positive rate (TPR) as x and y axes respectively. Experimental results show that our approach outperforms the traditional one which is based on visual features, and achieves satisfactory performance. The true positive rate achieves 96.7% while the false positive rate is about 10%. Its performance outperforms many state-of-the-art methods, such as [33, 34, 36, 39-44].

## 4.1 Evaluation of our hybrid ROI detection model

We collected adult images from the Internet and divided them into three groups, each of which includes 350 images. Our experiments were comprised of four parts to capture salient regions, skin-color regions, no-face regions, and final results of ROI respectively. Due to the subjectivity of human attention perception, there is not a standardized objective correctness measure for image attention analysis evaluation. Therefore, we adopted the evaluation method proposed in Ma's paper [11]. Twenty non-professional persons were invited to assign one of assessments, GOOD, ACCEPT, or FAILED, to our experimental results.

**Table 1. Correctness assessments of our hybrid ROI detection model**

| Image group | Good | Accept | Failed |
|---|---|---|---|
| Group 1 | 41.71% | 49.43% | 8.86% |
| Group 2 | 41.43% | 51.14% | 7.43% |
| Group 3 | 46.29% | 44.00% | 9.71% |
| **Average** | 91.33% | | 8.67% |

Table 1 lists the evaluation results of all the ROI. According to the evaluation results, we can conclude that our hybrid ROI detection model achieves good performance and the precision reaches 91.33% in average. Furthermore, we compare the salient regions' area among Itti's model, Ma's model, and our ROI model, as shown in Figure 4. Consequently, we can conclude that our ROI model is better than Itti's model [10] and Ma's model [11] from the point of the views of "good detection" and "high resolution".

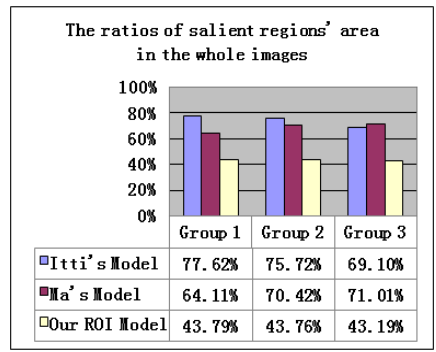

**Fig.4. Comparisons of salient regions' area among Itti's model, Ma's model, and our ROI model**

## 4.2 Evaluating saliency based visual codebook algorithm

In Figure 5, SIFT-Whole, SIFT-Skin, and SURF-ROI are ROC curves respectively on behalf of the traditional codebook algorithm (SIFT-Whole) [37-38], the codebook algorithm based on SIFT and skin-region (SIFT-Skin) [40], and saliency based visual codebook algorithm (SURF-ROI). Results show that saliency based visual codebook algorithm can remarkably improve the performance.

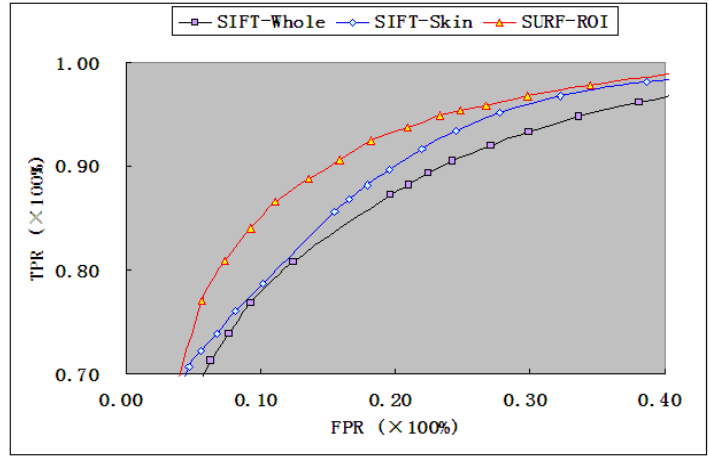

**Fig.5. Performance comparisons of visual codebook representation**

### 4.3 Evaluating periodicity based audio codebook algorithm

Audio signals are distilled from videos and shown in the format of ".wav". The training and testing videos are respectively segmented into 60941 and 69211 EEs on the basis periodicity analysis. Some parameters, for instance the consecutive counter number N, are respectively regulated. And their ROC curves are given in Figure 6.

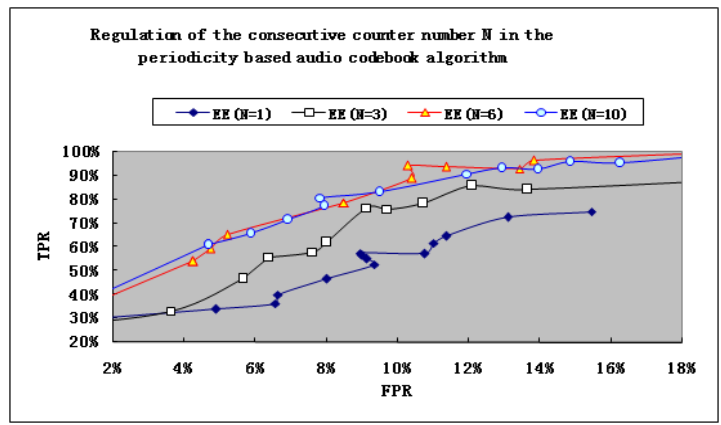

**Fig.6. Regulating parameters in periodicity based audio codebook algorithm**

### 4.4 Evaluating the performance of our approach

Experimental results show that our approach outperforms the traditional one which is based on visual features, and achieves satisfactory performance. As shown in Figure 7, the true positive rate (TPR) achieves 96.7% while the false positive rate (FPR) is about 10%. Its performance outperforms many existing methods, such as the literatures [33, 34, 36, 39-44].

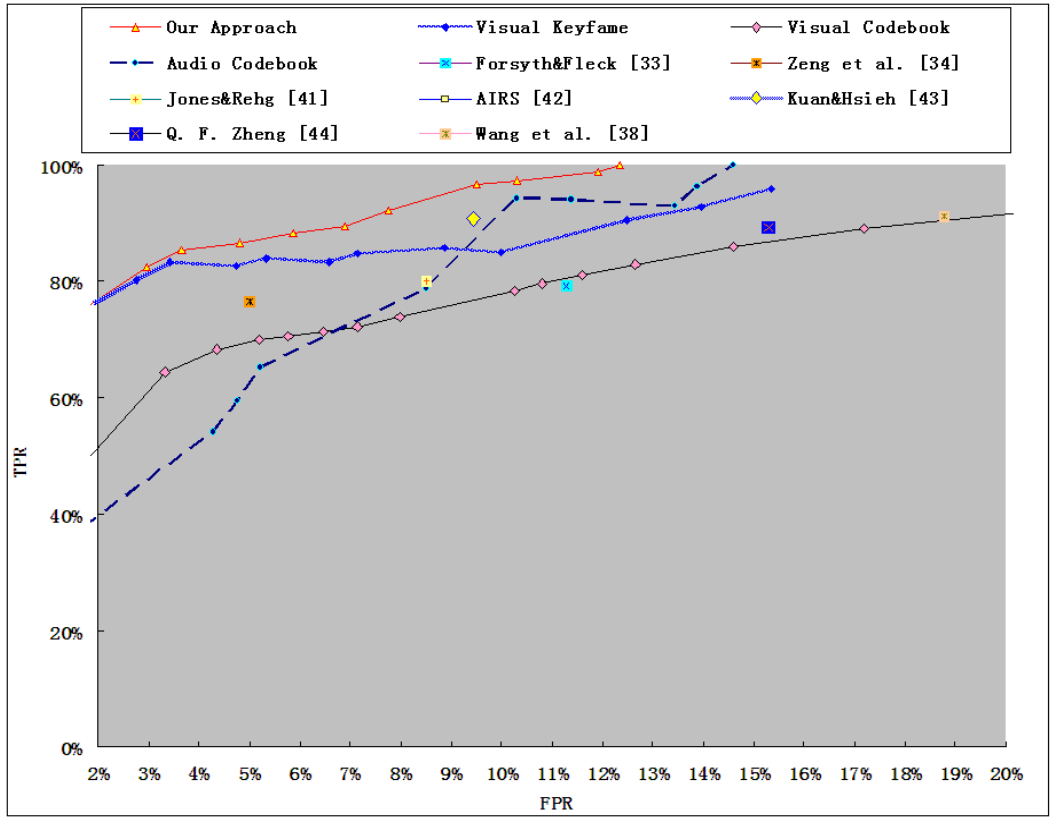

**Fig.7. Performance comparisons among some state-of-the-art methods**

# 5 Conclusions

Multi-modality is an effective approach of filtering pornography in adult video detection. But the performance of existing methods is not good enough due to the lack of accurate multi-modality semantics representation. Therefore, we present a novel approach of analyzing periodicity and saliency for adult video detection. We successively illustrate audio codebook algorithm and visual codebook algorithm which is respectively based on periodicity and saliency. Then they are combined to represent multi-modal co-occurrence semantics. Moreover, the performance of our approach is evaluated step by step. Experimental results show that our approach achieves satisfactory performance and outperforms the traditional one which is based on visual features. The true positive rate achieves 96.7% while the false positive rate is about 10%. Its performance outperforms many state-of-the-art methods. In this paper, we focus on multimodal semantic representation of videos. The main defect of the proposed method is its computational performance. We intend to address this issue in our future work.

***Acknowledgments***. This work is supported by National Nature Science Foundation of China (61702179); Hunan Provincial Natural Science Foundation of China (2018JJ4052, 2017JJ2099, 2017JJ2081, 2017JJ3091).